# A Study on Clustering for Clustering Based Image De-Noising

Hossein Bakhshi Golestani*
Department of Electrical Engineering, Sharif University of Technology, Tehran, Iran
h.b.golestani@gmail.com
Mohsen Joneidi
Department of Electrical Engineering and Computer Science, University of Central Florida, Orlando, USA
Joneidi@knights.ucf.edu
Mostafa Sadeghii
Department of Electrical Engineering, Sharif University of Technology, Tehran, Iran
m.saadeghii@gmail.com

**Abstract**
In this paper, the problem of de-noising of an image contaminated with Additive White Gaussian Noise (AWGN) is studied. This subject is an open problem in signal processing for more than 50 years. Local methods suggested in recent years, have obtained better results than global methods. However by more intelligent training in such a way that first, important data is more effective for training, second, clustering in such way that training blocks lie in low-rank subspaces, we can design a dictionary applicable for image de-noising and obtain results near the state of the art local methods. In the present paper, we suggest a method based on global clustering of image constructing blocks. As the type of clustering plays an important role in clustering-based de-noising methods, we address two questions about the clustering. The first, which parts of the data should be considered for clustering? and the second, what data clustering method is suitable for de-noising.? Then clustering is exploited to learn an over complete dictionary. By obtaining sparse decomposition of the noisy image blocks in terms of the dictionary atoms, the de-noised version is achieved. In addition to our framework, 7 popular dictionary learning methods are simulated and compared. The results are compared based on two major factors: (1) de-noising performance and (2) execution time. Experimental results show that our dictionary learning framework outperforms its competitors in terms of both factors.

**Keywords:** Image De-Noising; Data Clustering; Dictionary Learning; Histogram Equalization and Sparse Representation.

## 1. Introduction

We consider the problem of estimating a clean version of an image contaminated with Additive White Gaussian Noise (AWGN). A general approach to this aim is division of the noisy image into some (overlapping) small blocks, then de-noising of each block and finally obtaining the overall estimation of the clean image by averaging the de-noised blocks. The model is as follows:

$$y_i = z_i + n_i \qquad (1)$$

where $y_i$ is the vector form of the $i$th block of the noisy image, $z_i$ is the vector form of the $i$th block of the original image, and $n_i$ is a zero-mean AWGN with variance $\sigma^2$. Throughout the paper, the blocks are $n \times n$, thus the vector space dimension is $n^2$.

Image de-noising is still an open problem and numerous methods have been suggested up to now. The methods are based on defining a neighborhood for each block and weighted averaging according to suitable weights. The weights are computed in each neighborhood, as in [1-4] which are some relatively successful approaches. All of them are in the spatial domain. The method in [5] can be considered as same as [1-4], where processing is conducted in frequency domain. This method constructs a three-dimensional matrix by grouping those blocks that are similar (in some senses, e.g. $\ell_2$ norm) with a block of the image. Corresponding to each block of the image a group of similar blocks should be found. In this way, a three-dimensional matrix is obtained corresponding to each block. Then, a 3D collaborative signal filtering in the frequency domain is performed which provide a good estimation of the clean version of each block. This method can be considered as the state of the art method of image de-noising; however it suffers from high computational complexity due to local processing. The work in [6] has the same approach and applied filtering in the Principal Component Analysis (PCA) transform domain. Elad and Aharon [7] have suggested a new approach. They have used K-Singular Value Decomposition (K-SVD), which is a dictionary learning algorithm, to produce a global dictionary using the noisy image blocks. This method uses the representation in terms of the dictionary to de-noise image. The estimate of each de-noised block can be estimated by analyzing noisy blocks in this dictionary and applying a sparse recovery algorithm.

Local and global methods have some advantages and disadvantages. A global dictionary can recover general characteristics of an image, which are repeated in its several regions. However, these methods are not able to recover special local textures and details in an image.

While local methods indicate higher efficiency in recovering local details of image, they encounter over-learning risk leading from noise learning and incorporating noise into the final result. Deficiency of learning in some regions is another problem of local methods.

In [8], a clustering-based method was suggested. This method produces a local dictionary by clustering feature vectors from all noisy image blocks and conducts de-noising using decomposition of noisy blocks in terms of representatives of the found clusters. Similar to K-SVD, this method is based on dictionary but it uses a local dictionary.

Local patching and similar blocks clustering are effective factors in success of methods including [5], [6] and [8]. Dictionary learning based de-noising methods also perform some type of blocks clustering, for example K-SVD is a generalization of K-means clustering algorithm. So it is necessary to consider the clustering for the de-noising application more closely.

In this paper, we propose an approach for constructing a global dictionary and de-noising based on sparse decomposition of noisy blocks over the dictionary. This global dictionary is constructed by aid of the optimized clustering that will be presented. In the following sections clustering of image blocks is studied with more details in section 2. An analytical comparison between local and global clustering is addressed in section3. Section 4 studies the effect of equalization of data according to their variance in order to have an appropriate clustering. Learning the dictionary is explained in section 5 based on representatives of the found clusters. Section 6 studies applying of de-noising using dictionary. Finally, the local and global methods are evaluated in section 7.

## 2. Clustering of Image Blocks

In the case of methods including LPG-PCA, KLLD, BM3D ([6], [8] and [5], respectively), grouping of similar blocks is their critical factor of success. So, blocks grouping may has details which should be considered specifically. BM3D and LPG-PCA perform de-noising by clustering of the set of image blocks. K-LLD method performs clustering on feature vector extracted from surrounding blocks (Corresponding to each block). Considering the number of pixels and feature vector dimension, this clustering is of high computational load. In addition to high computational load, unbalanced clustering is one of the problems of global clustering of blocks. This problem is shown in Figure 1.

Assume that in Figure 1-bottom, the goal is to find 2 means. K-means algorithm finds two datacenters indicated by violet circles. These points are not good representatives of the blocks corresponding to the image edges. However, clustering objective function is minimized by these centers. Dense (high number data) correspond to image smooth parts and scattered (low number data) correspond to blocks containing edge or special texture. Traditional clustering algorithms behave with data corresponding to high energy areas as outlier data. So, these blocks have minor effect on the training by common clustering methods and the final desirable result will not be obtained. To solve the problem, first limitations of clustering-based de-noising methods are examined.

The MSE error lower bounds for image de-noising have been examined in [9] and [10]. This lower bound for one n × ncluster block is calculated as follow.

$$E[\|z_i - \hat{z}_i\|^2] \geq Trace[(J_i + C_z^{-1})^{-1}] \quad (2)$$
$$C_z = C_y - \sigma^2 I \quad (3)$$

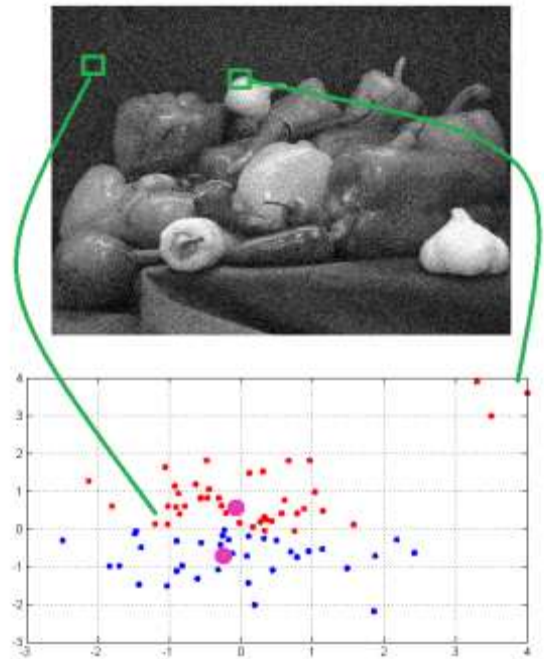

Fig 1. In natural images, number of smooth blocks are more than high energy ones.

where, $J_i$ is the Fisher information matrix and $C_z$ is the estimated covariance matrix for the group of vectors that are similar to $i$th block. For zero mean Gaussian noise, [10] assumed matrix $J_i$ as follow:

$$J_i = \frac{N_i}{\sigma^2} I \quad (4)$$

where, $N_i$ is the number of similar vectors of the $i$th block. Assuming that similar vectors for each pixel are of many members and noise level is not high, the right hand of inequality is simplified:

$$(J_i + \hat{C}_z^{-1})^{-1} \cong \frac{\sigma^2}{N_i}\left(I - \frac{\sigma^2}{N_i}\hat{C}_z^{-1}\right) \quad (5)$$

$$E[\|\mathbf{z}_i - \hat{\mathbf{z}}_i\|^2] \geq \frac{\sigma^2}{N_i} Trace(I - \frac{\sigma^2}{N_i}\hat{C}_z^{-1}) \quad (6)$$

$$E[\|z_i - \hat{z}_i\|^2] \geq \frac{\sigma^2}{N_i}(n^2 - \frac{1}{N_i}\sum_{j=1}^{n^2}\frac{\sigma^2}{\lambda_j}) \quad (7)$$

where $\lambda_j$ is the $j$th eigenvalue of covariance matrix of estimated data $\hat{C}_z$:

$$\lambda = eig(\hat{C}_z) = eig(C_y) - \sigma^2 \quad (8)$$

Assuming that the number of similar patches of each block and the noise level is the same for all blocks; thus de-noising bound is related to covariance matrix. High detailed clusters (having high covariance matrix eigenvalues) are more difficult to de-noise. So for blocks corresponding to low complex areas, lower bound will be decreased for MSE of the estimated version and the original image. However the result is predictable; because in smooth areas of an image, a simple averaging can obtain good result but if a block consists of more complexity, specific texture and high variance, would limit de-noising performance. For such blocks, more precise similar block grouping is needed. The more the number of same blocks causes the more appropriate characteristics of grouping. So we suggest that for detailed and textured blocks, more training data should be used.

Let us generalize the concept presented in (2) to clusters (rather than groups for each block). Assume variable i is allocated for clusters rather than blocks in (2). In other words, $z_i$ is a block from the ith cluster and $N_i$ is the number of members of the $i$th cluster. $C_z$ is the estimated covariance matrix of the $i$th cluster.

First question that this paper is going to answer is "which blocks should be considered for clustering?" As stated before, using all blocks for clustering not only have high computational load but also leads to unbalanced clustering. Figures 4 and 5 illustrate the idea of equalized clustering. Figure6 is the equalized clustered of Figure 1 providing good properties for de-noising application. Dictionary learning-based methods such as K-SVD decrease training data in a random way to reduce computational load. But as have been seen, removing valuable blocks from training data has negative effect on the de-noising lower bound. In Figure 6, only data corresponding to smooth blocks are removed and the obtained cluster centers are more appropriate for de-noising. In section 3 training data equalization will be studied.

Second question that the paper is going to answer is "how do the clustering?" Now we state the problem of clustering. First we rewrite (2) as follow:

$$E[\|z_i - \hat{z}_i\|^2] \geq \frac{\sigma^2}{N_i}\sum_j \frac{\lambda_j}{\lambda_j + \frac{\sigma^2}{N_i}} \quad (9)$$

Let us write the right side of this inequality for all clusters as a cost function:

$$J(\Omega) = \sum_i \frac{\sigma^2}{N_i}\sum_j \frac{\lambda_j}{\lambda_j + \frac{\sigma^2}{N_i}} \quad (10)$$

$\Omega$ is the set of indices of training data that shows membership of the training data to clusters. The problem of the optimum clustering can be stated as follows:

$$\min_\Omega J(\Omega) = \sum_i \frac{\sigma^2}{N_i}\sum_j \frac{\lambda_{ij}}{\lambda_{ij} + \frac{\sigma^2}{N_i}} \quad (11)$$

The above problem is dependent of Eigenvalues of each cluster $\lambda_{ij}$, so its computational burden is very high. Thus, exact solution of the problem is not achievable. Eigen values of the clusters corresponding to smooth or constant regions of $\hat{z}$ are about zero so they can be neglected from $J(\Omega)$. So, only high variance blocks affect the cost function.

$$J(\Omega) \cong \sum_{\substack{none\,smooth \\ clusters}} \frac{\sigma^2}{N_i}\sum_j \frac{\lambda_j}{\lambda_j + \frac{\sigma^2}{N_i}} \quad (12)$$

In other words, smooth training data can be ignored in the clustering. At the first glance this simplification just makes the clustering fast but it has an effect on the accuracy of the clustering. In fact, less exploitation of non-important blocks causes in more affection of important blocks in the clustering problem (compare figure 1 and figure 6). Eq. (12) can be interpreted as a hard threshold for selection of blocks in clustering. In the next section variance of blocks will be introduced as a criterion for smoothness and then variance histogram equalization will be presented as the soft threshold version of (12) for selection of data that participate in clustering.

Problem (11) can be viewed from another point of view. The cost function encourages clusters to have a sparse vector of Eigen values. Figure 2 shows how (11) encourages Eigen values to be zero. In other words problem (11) clusters data into low-rank subspaces and guarantees that many of Eigen values will be zero for each cluster.

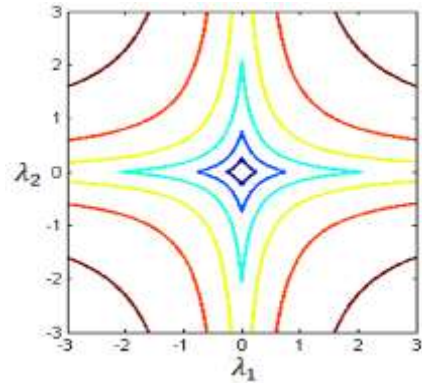

Fig 2. Contour of cost function of (11) for a cluster[1].

---

[1] The figure is contour of $\sum_j \frac{|\lambda_j|}{|\lambda_j|+\frac{\sigma^2}{N_i}}$, as values of $\lambda$ are positive, figure 2 is true for contour of (11)

High dimensional data that lie in low-rank subspaces have high correlation with each other (see Figure 3). An alternative for subspace clustering may be correlation clustering [11] that has much less computational load. As can be seen in Figure 3, the obtained clusters by correlation clustering lie in a rank-1 subspace that agrees with problem (11) because only one Eigen value of the covariance matrix of this cluster is none-zero. In section 6 simulations has been done by correlation clustering.

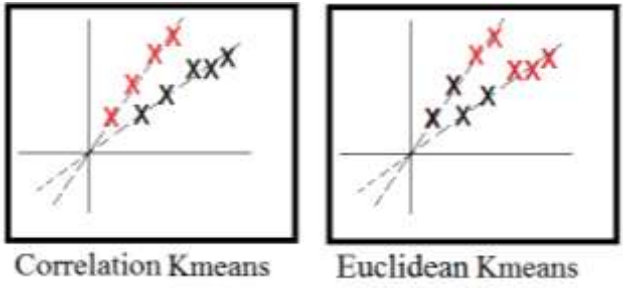

Fig 3. Comparison of correlation clustering and traditional clustering.

## 3. Global Clustering vs. Local Clustering

A well-known clustering method is the family of K-means clustering algorithms [12], which have been used by K-LLD [8] for image de-noising. K-means clustering algorithm solves the following problem

$$\min_{D} \sum_{k=1}^{K} \sum_{j \in \Omega_k} \| y_j - d_k \|_2^2 \qquad (13)$$

where, $D = [d_1, \ldots, d_k]$. This problem can be written in the following form which is a factorization

$$\min_{D,X} \| Y - DX \|_F^2 , \forall i,j : \| x_i \|_0 = 1, x_i^j \in \{0,1\} \qquad (14)$$

where, Y=$[Y_1, \ldots, Y_L]$ ($L$ is the number of blocks), $x_i$ is the $i$th column of $X$, and $x_i^j$ is the $j$th entry of $x_i$. This problem implies that all entries of each $x_i$ must be equal to zero except one of them. The non-zero element is forced to be 1. This restriction does not exist in the so-called gain-shaped variant of K-means [12], which solves the following problem

$$\min_{D,X} \| Y - DX \|_F^2 \; subject \; to \; \forall i : \| x_i \|_0 = 1 \qquad (15)$$

This problem is a K-rank1 subspace (K-lines) clustering. As can be seen in Fig. 4 (b) and (d), the obtained clusters by gain-shaped K-means is in agreement with problem (11). This is because only one eigenvalue of each cluster's covariance matrix is non-zero.

Inspired by the simple approach (15), a suboptimal solution for (11) can be obtained. We propose to construct the proper basis using the obtained cluster centroids and dominant principal components (PCs) of each cluster (generally, natural images are not perfectly lie on rank-1 subspace as in Fig. 4, i.e., thus the proposed dictionary also contains dominant PCs spanning details of each cluster). Those PCs would be added to the dictionary if their corresponding eigenvalues are greater than noise variance. The noisy image blocks are then de-noised inspired by the framework used in [7]. This leads to a fast and efficient de-noising algorithm (algorithm1). It will be shown in Section 7 that the proposed algorithm outperforms traditional K-SVD.

**Algorithm 1** Image de-noising based on gain-shaped K-means
1: **Task** De-noise **Z** from AWGN with variance $\sigma^2$
2: Learning $K$ columns of **D** using K-subspace [13]
3: Construct the dictionary by the cluster centroids and most significant PCs
4: **for** $i = 1, \ldots, L$ **do**
5:     Sparse code $y_i$
6:     Estimate $z_i$ by projection on the dictionary
7: **end for**
8: Construct the de-noised image by $\{z_i\}_{i=1}^{L}$

Another approach for clustering is *dictionary learning* in sparse signal representation, which aims to solve the following problem

$$\min_{D,X} \| Y - DX \|_F^2 \; subject \; to \; \forall i : \| x_i \|_0 \leq \tau \qquad (16)$$

K-SVD is a well-known dictionary learning algorithm. Low-rank subspaces found by K-SVD have overlaps. It means that corresponding to each subset of the columns of $D$, there is a low-rank subspace that K-SVD learns. Data that used the same subset lie on a low-rank subspace but K-SVD learns a very large number of low-rank subspaces for a set of training data such that many of them are empty or low populated (refer to Fig. 5, top). Actually, clusters found by K-SVD include the data that have used the same dictionary columns. Note that these clusters are not guaranteed to be low-rank. In the simulation results we will see that our proposed method based on gain-shaped K-means outperforms K-SVD.

The derived problem (11) describes a suitable global clustering problem, while the state of the art algorithms do not perform global clustering, but instead use local patch-grouping. Translating global clustering to local grouping converts the problem to,

$$G_i = \min_{G} \| \lambda_G \|_0 \; subject \; to \; |G| \geq \tau, G \in W_i, i \in G \qquad (17)$$

where, $G_i$ is group of blocks corresponding to the $i$th block, $\lambda_G$ is the eigenvalues of covariance matrix of $G_i$ and $W_i$ is a window around the $i$th block. The last constraint implies that the $i$th block must be member of $G_i$. An equivalent form of (17) can be stated as,

$$G_i = \max_{G} |G| \; subject \; to \; \| \lambda_G \|_0 \leq \tau, G \in W_i, i \in G \qquad (18)$$

BM3D, a high performance image de-noising algorithm, implicitly uses (18) in order to perform local grouping. The similarity criterion used in BM3D for performing local grouping is novel, in which firstly blocks are transformed using an orthonormal transformation (e.g., DCT and DFT), then a projection on a low-rank subspace is performed using hard-thresholding of the coefficients of each block. In the new transformed space, a simple Euclidean distance determines similar blocks with the *i*th block. Truncated coefficients of the similar blocks with the *i*th one also lie on a low-rank subspaces near to the *i*th one, thus many of $\lambda_{G_i}$ are about zero and the constraint of (18) is satisfied.

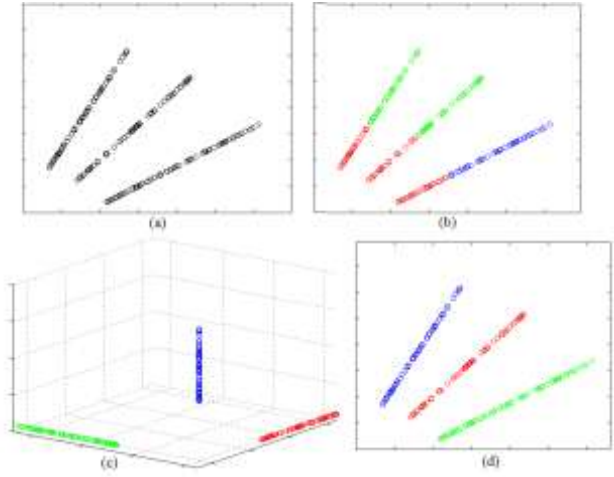

Fig 4. Comparison of clustering in raw data domain and in the sparse-domain transformed data (as used in CSR and LSSC) for some 2D data. (a) Raw data. (b) K-means clustering on raw data (K=3). (c) K-means clustering on sparse-domain transformed data using an over-complete dictionary having 3 atoms. (d) Reconstruction of the data from their sparse representations in (c), in the case of these data Gain-shaped K-means directly results in (d).

The idea behind (18) can be used in another way different from what BM3D has used. These de-noising algorithms first perform grouping using a rough criterion, e.g. Euclidean distance, then in the main de-noising algorithm obtain a low-rank representative for each group and use it. The algorithm suggested by Dong *et al*. (clustering based sparse representation or CSR) [13] which solves the following problem, is an example of these types of algorithms

$$\min_{X,B} \| Y - DX \|_F^2 + \gamma_1 \sum_i \| x_i \|_0 + \gamma_2 \sum_{k=1}^{K} \sum_{j \in G_k} \| x_j - b_k \|_2^2 \quad (19)$$

where $= [b_k]$, and $b_k$ is the centroid of the *k*th group. Note that (19) does not optimize the dictionary. In fact, firstly a global dictionary using K-means and PCA is learned which is then used by this problem to simultaneously perform local grouping and sparse coding, in an iterative procedure. The first and second terms in (19) are similar to K-SVD problem, but the last term clusters the sparse-domain transformed data. Figure 5 illustrates the effect of clustering data in the sparse domain rather than the raw data. Contrary to K-SVD, in which the members of a cluster have used one column of *D*, problem (19) encourages the clustering to put data that have the same sparse representation (structure) in one cluster.

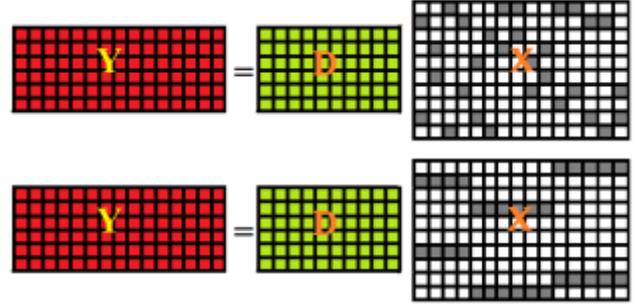

Fig 5: Top: K-SVD approximates data by a union of rank-2 subspaces. No rank-2 cluster can be found. Bottom: Group sparsity constraint on *X*. There are three rank-2 clusters.

Another local grouping based method is a novel approach, called learned simultaneous sparse coding (LSSC) [14], that simultaneously performs group sparse coding [15] and grouping the similar patches. Group sparse coding implies that the blocks within a group have similar sparse representations, like CSR. This is achieved by jointly decomposing groups of similar signals on subsets of the learned dictionary (as previously explained, K-SVD fails to achieve this goal. See Fig. 5 for comparison). They proposed the following cost function,

$$\min_{X_k} \sum_{k=1}^{K} \| X_k \|_{p,q} \; s.t. \; \forall k : \sum_{i \in G_k} \| y_i - Dx_i^k \|_2 \leq \varepsilon \quad (20)$$

where, $X_k$ is the coefficient matrix of the *k*th cluster data, $X_j^k$ is the *j*th column of $X_k$, and $\|X\|_{p,q} = \sum_i \|x_{[i]}\|_q^p$, with $x_{[i]}$ the *i*th row of *X*. Minimizing $\|X\|_{p,q}$ with *p*=1 and *q*=2 (that is, the $\ell_1$ norm of the vector containing the $\ell_2$ norms of the rows) implies that the number of engaged rows of *X* will be limited. In other words, this cost function encourages the data to have the same support of sparse representation in a cluster. As the data in the same cluster can be decomposed by few bases, the rank of the data matrix in the same cluster will be minimized. Thus a solution for (20) tries to minimize (17). i.e, $\Sigma\|X_k\|_{p,q}$ approximates $\|\lambda_{Gk}\|_0$. At the simulation results section, numerical performances of the explained local and global methods are compared, separately.

## 4. Block Variance Histogram Equalization

For the reasons previously stated some points should be considered. Firstly clusters with different complexities have approximately the same number of members. Secondly, members of complicated clusters should not have high distance from cluster subspace so that covariance matrix eigenvalues would not become high and many of them would be zero. Third, members of high complex clusters should not be neglected for dictionary learning.

Blocks variance is considered as a complexity measure. In natural images, the number of high complex blocks is lower than low complex blocks. Figure 6 indicates blocks variance histogram of an original image and its noisy version. As can be seen, in the original image, concentration is in lower values of variance and in noisy image concentration is in the point corresponding to noise variance representing smooth blocks of original image. Those blocks that their variances are approximately the same as noise variance are not useful for training. Using these blocks not only increases computational load but also causes unbalance clustering and reduces the effect of important clusters. So their number in final clustering should be reduced. To equalize blocks variance histogram, an equalization transform function must be used. The following function is an example:

$$T(\sigma) \triangleq \begin{cases} \frac{th}{p(\sigma)} & p(\sigma) > th \\ 1 & p(\sigma) < th \end{cases} \quad (21)$$

where, $p(\sigma)$ is density function of blocks variance probability and th is a threshold. $T(\sigma)$ is the probability of entering a block with variance $\sigma$ into training data to be used for clustering. Figure 6, indicates an example of the transform function and equalized histogram of noisy image in Figure 7. In this histogram, the effect of blocks with variance 25 is reduced considerably. Figure 8 shows equalized clustering of figure 1.

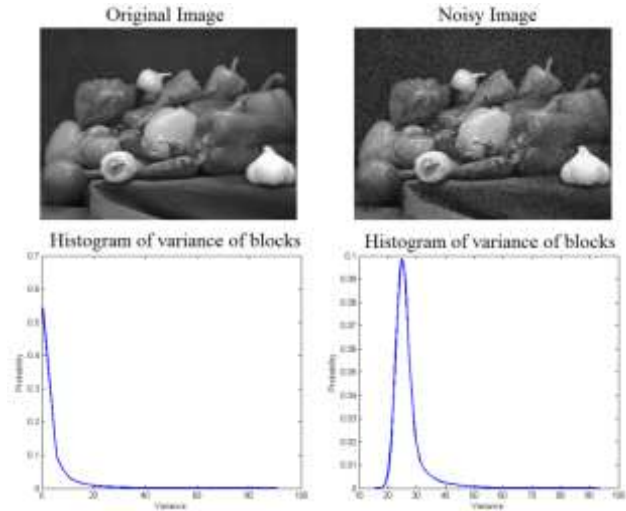

Fig 6. two clear and noisy images with $\sigma = 25$ and their blocks variance histogram.

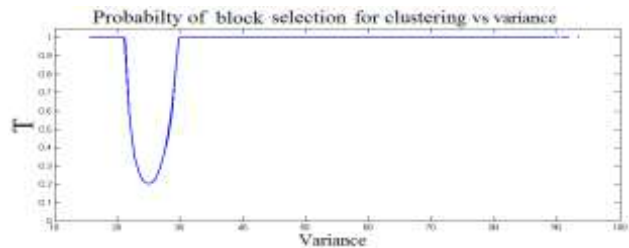

Fig 7. Equalizing transform function

Regarding that the variance of smooth blocks is approximately the same as noise variance. It can be said that there is not valuable information about original image, and their presence for training not only mislead the clustering algorithm but also have high computational load. Now, subspace clustering should be done on remaining training data which agrees with Eq. (11).

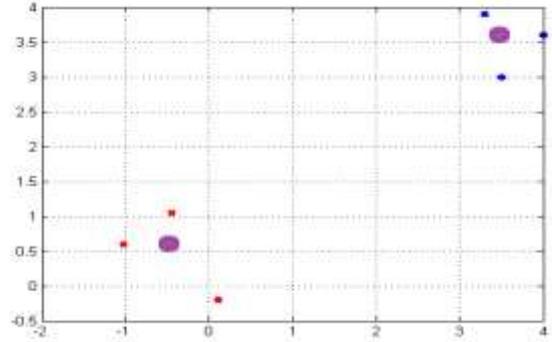

Fig 8. Equalized clustering of figure 1

## 5. Dictionary Learning

Dictionary learning is performed using the blocks selected in the previous stage. The final dictionary includes $P_i$ dominant principal components from each cluster (equal to non-zero Eigen values of matrix $\hat{C}_z$ explained in Section 2).

In the next stage, SVD transform is derived from covariance of data matrix of each cluster:

$$Y_i Y_i^T = U_i \Lambda_i V_i^T, i = 1 \dots K \quad (22)$$

where, $K$ is the number of clusters. Singular values on the main diagonal $\Lambda_i$ are equal to $\lambda_{ij}$ which are arranged in ascending order by $j$ subscript.

For each cluster, $P_i$ is the number of principal components that will be included in the final dictionary and is obtained by the following equation:

$$P_i = \left( \underset{j}{argmax} \ \lambda_{ij} \ \middle| \ \lambda_{ij} \leq \sigma^2 \right) - 1$$
$$U_i(:,1:P_i) \in D \quad (23)$$

The principal components higher than $P_i$ have learned noise for each cluster in matrix $U_i$. Actually, $P_i$ is the dimension of noise-free data on the $i$th cluster (or $P_i$ is the rank of subspace that ith cluster lies in it). It means that if the noise power is zero, autocorrelation matrix of $i$th cluster has only $P_i$ non-zero eigenvalues. In presence of noise, all autocorrelation matrix eigenvalues of each cluster of noisy data will be nonzero; from the component $P_i + 1$ to the end are due to noise. By adding the first principal component to $P_i$, the dictionary is completed and we can perform denoising by this designed dictionary.

## 6. Denoising Operation

Usefulness of the union of subspaces model has been proved in many applications of signal processing. As illustrated in section 2 and 3, this model is appropriate for the analysis of signal de-noising. This model assumes that image blocks are linear combination of few bases of a dictionary:

$$z_i = D\alpha_i \; st \; \|\alpha_i\|_0 \leq Th \tag{24}$$

In the previous section a dictionary was defined. De-noised image should also meet this model whereas noisy image $y_i$ cannot, because in the dictionary learning stage, noise is not trained. In other words, to represent noise, many bases combination should be involved and no sparse representation $\alpha_j$ in equation (25) can be found.

$$\nexists \; \alpha_i \; | \; y_i = D\alpha_i \; st \; \|\alpha_i\|_0 \leq Th \tag{25}$$

The model must be reformed to model the noise of data:

$$y_i = D\alpha_i + n_i \; , \|\alpha_i\|_0 \leq Th \tag{26}$$

Assuming Gaussian noise with zero mean in this model, MAP estimation for $\alpha_j$ is

$$\hat{\alpha}_i = \min_{\alpha_i} \|y_i - D\alpha_i\|^2 \; st: \|\alpha_i\|_0 \leq Th \tag{27}$$

Optimum threshold is related to $P_i$ of a cluster where $y_i$ belongs to it. This can be replaced by the following problem:

$$\hat{\alpha}_i = \min_{\alpha_i} \|\alpha_i\|_0 \; st: \|y_i - D\alpha_i\|^2 \leq \epsilon \tag{28}$$

where, $\epsilon$ is a function of noise variance. Now we can estimate de-noised version by this estimation of sparse coefficients. We just need to project $y_i$ into the nearest low-rank subspace spanned by the columns of the learned dictionary.

## 7. Simulation Results

In this section, de-noising results of proposed method and some other recent approaches are presented and discussed. First, the global and local methods are evaluated, then a comparison between global and local approaches is presented and finally these methods are compared in term of total execution time.

K-SVD and our simple gain-shaped K-means (proposed method) are presented as global methods. The presented local methods include those introduced in [5], [8], [13], [14], [17] and [18]. Performance comparison of these algorithms can be seen Table 1. We have used the Peak Signal to Noise Ratio (PSNR[1]) as the performance criterion. The PSNR values were averaged over 5 experiments, corresponding to 5 different realizations of AWGN. The variance was negligible and not reported.

Our method is simulated similar to the framework of [7]. Both algorithms have the same amount of error for the training set (depending on the noise variance) but their size of dictionary is different. Table 1 shows that the proposed method surpasses the K-SVD [7] and its results are comparable with the time consuming local methods. As will be tabulated, the execution time of the proposed method is about 70% of K-SVD, 8% of LSSC [14] and 4% of CSR [13]. Recently [16] investigated a comprehensive comparison of different image de-noising methods. They have shown numerically that BM3D, SCR and LSSC studied in this paper have the best results. Figure 9 shows an example of de-noising results by our proposed method.

Table 1. Image de-noising performance of the Global and Local methods in PSNR (dB) for 4 different image and various $\sigma/SNR$

| | | Lena | | |
|---|---|---|---|---|
| | $\sigma/SNR$ | 5/34.16 | 10/28.14 | 20/22.11 |
| Global | Proposed | 38.71 | 35.60 | 32.57 |
| | K-SVD [7] | 38.60 | 35.47 | 32.38 |
| Local | K-LLD [8] | 38.01 | 35.20 | 32.37 |
| | LSSC [14] | 38.69 | 35.83 | 32.90 |
| | CSR [13] | 38.74 | 35.90 | 32.96 |
| | BM3D [5] | 38.72 | 35.93 | 33.05 |
| | LSC [17] | 38.56 | 35.65 | 32.54 |
| | SSMS [18] | 38.62 | 35.63 | 32.30 |
| | | Barbara | | |
| | $\sigma/SNR$ | 5/34.16 | 10/28.14 | 20/22.11 |
| Global | Proposed | 38.22 | 34.68 | 30.98 |
| | K-SVD [7] | 38.08 | 34.42 | 30.83 |
| Local | K-LLD [8] | 37.26 | 33.30 | 28.93 |
| | LSSC [14] | 38.48 | 34.97 | 31.57 |
| | CSR [13] | 38.43 | 35.10 | 31.78 |
| | BM3D [5] | 38.31 | 34.98 | 31.75 |
| | LSC [17] | 38.45 | 34.95 | 31.29 |
| | SSMS [18] | 38.73 | 35.11 | 31.25 |
| | | House | | |
| | $\sigma/SNR$ | 5/34.16 | 10/28.14 | 20/22.11 |
| Global | Proposed | 39.59 | 36.54 | 33.68 |
| | K-SVD [7] | 39.37 | 35.98 | 33.20 |
| Local | K-LLD [8] | 37.63 | 35.09 | 32.66 |
| | LSSC [14] | 39.93 | 36.96 | 34.16 |
| | CSR [13] | 39.98 | 36.88 | 33.86 |
| | BM3D [5] | 39.83 | 36.71 | 33.77 |
| | LSC [17] | 39.72 | 36.33 | 33.23 |
| | SSMS [18] | 39.51 | 36.13 | 32.77 |
| | | Boat | | |
| | $\sigma/SNR$ | 5/34.16 | 10/28.14 | 20/22.11 |
| Global | Proposed | 37.25 | 33.85 | 30.52 |
| | K-SVD [7] | 37.22 | 33.64 | 30.36 |
| Local | K-LLD [8] | 35.96 | 33.16 | 30.17 |
| | LSSC [14] | 37.35 | 34.02 | 30.89 |
| | CSR [13] | 37.31 | 33.88 | 30.78 |
| | BM3D [5] | 37.28 | 33.92 | 30.87 |
| | LSC [17] | 37.16 | 33.75 | 30.42 |
| | SSMS [18] | 37.09 | 33.70 | 30.40 |

---

[1] PSNR is defined as $10log_{10}(255^2/MSE)$ and measured in dB

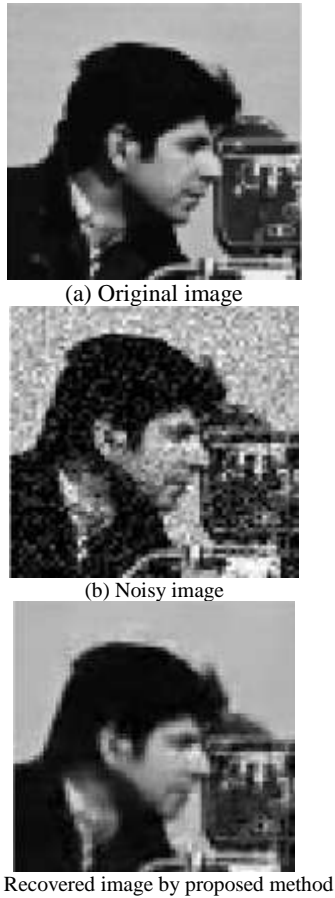

(a) Original image

(b) Noisy image

(c) Recovered image by proposed method

Fig 9. an example of denoising results by our method

In natural images, far away block have generally different patterns, so, using all blocks may result in inappropriate clustering. Moreover, non-overlapped clusters obtained by global methods are not as flexible as the overlapped groups. On the other hand, local grouping assign appropriate groups to each block. Although local methods have better performance, global methods are able to extract salient features of images and use it easily for de-nosing. According to comparison of local and global methods in Table 1, the performance of the proposed global method is just about $0.2dB$ lower than promising local methods (LSSC and CSR), which is not a high difference. However, a common good property of both global and local methods is that they exploit the low-dimensional characteristics of clusters/groups in order to design a suitable de-noising algorithm.

To understand the effect of this method on the dictionary, in the table 2 the results are compared only with K-SVD method, which is global a method like our proposed method. However, in the results of local methods in table 1, the suggested method used about 27% less blocks for training and the time required for dictionary learning is less than K-SVD method. This table studies the effect of data equalization on K-SVD. As it can be seen equalization improves K-SVD about 0.4dB.

Table 2. comparing the suggested method and K-SVD method. left: K-SVD + Equalization of data, middle: KSVD, right: the proposed clustering

| σ/SNR | House | | | Peppers | | |
|---|---|---|---|---|---|---|
| 20/22.11 | 33.29 | 33.16 | 33.68 | 30.89 | 30.77 | 31.09 |
| 25/20.18 | 32.37 | 32.19 | 32.66 | 29.82 | 29.69 | 29.96 |
| 30/18.59 | 31.40 | 31.24 | 31.61 | 28.95 | 28.82 | 29.11 |
| σ/SNR | Lena | | | Cameraman | | |
| 20/22.11 | 32.55 | 32.38 | 32.63 | 30.14 | 29.96 | 30.36 |
| 25/20.18 | 31.42 | 31.34 | 31.50 | 29.10 | 28.93 | 29.22 |
| 30/18.59 | 30.59 | 30.46 | 30.72 | 28.16 | 28.07 | 28.36 |

As mentioned, the proposed method is based on dictionary learning and its time efficiency should be compared with other dictionary learning based approaches e.g. [7], [13] and [14]. Table 4 compares the relative execution time of [13], [14], [7] and the proposed method in various image sizes. Our experiments were averaged on 5 different runs carried out on a Personal Computer with a 3.6-GHz AMD 2 Core CPU and 4 GB RAM. As can be seen, the global de-noising methods (KSVD and proposed) are more efficient in term of execution time and our proposed method surpasses KSVD. In fact, dictionary learning running time of proposed method (for identification of K-rank1 subspaces) is about 40% of K-SVD for 20,000 blocks extracted from a 512×512 image, but its overall execution time is about 72% of KSVD.

Table 3. Relative execution time of dictionary learning based methods (in minutes)

| Image Size | | 144×176 (QCIF) | 288×352 (CIF) | 576×704 (4CIF) |
|---|---|---|---|---|
| Local | LSSC [14] | 1.29 | 5.31 | 22.09 |
| | CSR [13] | 2.67 | 11.28 | 52.87 |
| Global | KSVD [7] | 0.14 | 0.59 | 2.66 |
| | Proposed | 0.10 | 0.43 | 1.92 |

## 8. Conclusions

Local methods suggested in recent years, have obtained better results than global methods. However by more intelligent training in such a way that first, important data is more effective for training, second, clustering in such way that training blocks lie in low-rank subspaces, we can design a dictionary applicable for image de-noising and obtain results near the state of the art local methods.

As was seen, we have obtained acceptable results by a relatively simple method based on construction of an appropriate global dictionary.

**Hossein Bakhshi Golestani** received the B.Sc. and M.Sc. in Electrical engineering from Ferdowsi university of Mashhad, Mashhad, Iran and Sharif University of Technology, Tehran, Iran, in 2010 and 2012 respectively. His research interests include multimedia signal processing, statistical signal processing, video/speech coding and compression, signal quality assessment and robotics.

**Mohsen Joneidi** received the B.Sc. and M.Sc. in Electrical engineering from Ferdowsi university of Mashhad, Mashhad, Iran and Sharif University of Technology, Tehran, Iran, in 2010 and 2012 respectively. He is currently working toward the Ph.D degree in department of electrical engineering and computer science at the University of Central Florida, Orlando, USA. His research interests include multimedia signal processing, statistical signal processing, signal quality assessment and compressive sensing.

**Mostafa Sadeghi** received the B.Sc. and M.Sc. in Electrical engineering from Ferdowsi university of Mashhad, Mashhad, Iran and Sharif University of Technology, Tehran, Iran, in 2010 and 2012 respectively. He is currently working toward the Ph.D degree in department of electrical engineering at the Sharif University of Technology, Tehran, Iran. His research interests include statistical signal processing, sparse representation/coding and compressive sensing.